\documentclass[journal,twoside,web]{ieeecolor}
\usepackage{tmi}
\usepackage{cite}
\usepackage{amsmath,amssymb,amsfonts}
\usepackage{algorithmic}
\usepackage{graphicx}
\usepackage{textcomp}
\usepackage[hypertexnames=false]{hyperref}
\usepackage[ruled,vlined,linesnumbered]{algorithm2e}
\usepackage{booktabs}
\usepackage{tabularx}
\usepackage{multirow}
\usepackage{multirow}
\usepackage{booktabs}
\usepackage[table]{xcolor}
\usepackage{colortbl}
\usepackage{booktabs}
\usepackage{adjustbox}
\usepackage{ulem}
\usepackage{afterpage}
\usepackage{float}
\usepackage{dirtytalk}
\usepackage{longtable}
\newcommand{\mbluetext}[1]{\textcolor{mblue}{#1}}
\usepackage{soul} 
\soulregister{\cite}7 
\soulregister{\citep}7 
\soulregister{\citet}7 
\soulregister{\ref}7 
\soulregister{\pageref}7 
\soulregister{\say}7 
\soulregister{\mbluetext}7 
\hypersetup{
    colorlinks=true,
    linkcolor=mblue,
    filecolor=magenta,      
    urlcolor=cyan,
}
\usepackage[switch]{lineno}

\def\BibTeX{{\rm B\kern-.05em{\sc i\kern-.025em b}\kern-.08em
    T\kern-.1667em\lower.7ex\hbox{E}\kern-.125emX}}
\begin{document}
\title{HAIBU-ReMUD: Reasoning Multimodal Ultrasound Dataset and Model Bridging to General Specific Domains}
\author{Shijie Wang, Yilun Zhang, Zeyu Lai, and Dexing Kong
\thanks{This work has been submitted to the IEEE for possible publication.
Copyright may be transferred without notice, after which this version
may no longer be accessible. 
(Corresponding authors:
Dexing Kong.)}
\thanks{Shijie Wang, Zeyu Lai, and Dexing Kong are with the School of Mathematical Sciences, Zhejiang University, Xihu, Hangzhou, Zhejang 310058,
China (e-mail: wang\_sj@zju.edu.cn; zyulai@zju.edu.cn; dxkong@zju.edu.cn).}
\thanks{Yilun Zhang is with Demetics Medical Technology and Prompt Technology, No.16, Zhejiang Overseas High-level Talents Innovation Park, Yu Hang District, Hangzhou, Zhejiang Province, China, 311121. e-mail: 3100105044@zju.edu.cn.}}

\maketitle

\begin{abstract}
Multimodal large language models (MLLMs) have shown great potential in general domains but perform poorly in some specific domains due to a lack of domain-specific data, such as image-text data or vedio-text data. In some specific domains, there is abundant graphic and textual data scattered around, but lacks standardized arrangement. In the field of medical ultrasound, there are ultrasonic diagnostic books, ultrasonic clinical guidelines, ultrasonic diagnostic reports, and so on. However, these ultrasonic materials are often saved in the forms of PDF, images, etc., and cannot be directly used for the training of MLLMs. This paper proposes a novel image-text reasoning supervised fine-tuning data generation pipeline to create specific domain quadruplets (image, question, thinking trace, and answer) from domain-specific materials. A medical ultrasound domain dataset ReMUD is established, containing over 45,000 reasoning and non-reasoning supervised fine-tuning Question Answering (QA) and Visual Question Answering (VQA) data. The ReMUD-7B model, fine-tuned on Qwen2.5-VL-7B-Instruct, outperforms general-domain MLLMs in medical ultrasound field. To facilitate research, the ReMUD dataset, data generation codebase, and ReMUD-7B parameters will be released at \mbluetext{\url{https://github.com/ShiDaizi/ReMUD}}, addressing the data shortage issue in specific domain MLLMs. 
\end{abstract}


\section{Introduction}
\label{sec:introduction}
\IEEEPARstart{I}{n} general domains, image-text data exists in abundance, like web images along with their corresponding captions and contexts. 
Benefit from this, multimodal large language models (MLLMs) instructions-tuned by leveraging multimodal inputs, such as LLaVA~\cite{liu2023visual} and GPT-4o~\cite{openai2024gpt4ocard}, have displayed remarkable zero-shot task completion performance across a wide range of user-oriented vision-language tasks, for instance, image understanding and reasoning. 
The rapid development of MLLMs has provided more possibilities for expansion into broader fields and application scenarios~\cite{song2025bridgegapmodalitiessurvey, liang2024survey, caffagni2024revolutionmultimodallargelanguage}.

Despite demonstrating excellent vision-language abilities in general domains, MLLMs' performance in specific domains often shows notable deficiencies~\cite{alsaad2024multimodal}. 
Training data generated by publicly available data usually lacks domain-specific knowledge, especially in medical domain, resulting in a lack of domain-specific expertise.  
General domain MLLMs may produce incorrect responses or complete hallucinations, which are not allowed to appear in medical domain~\cite{bai2024hallucination}.
Pure text-based large language models (LLMs) can be self-supervised trained through autoregression, predicting the next word.
However, in contrast, MLLMs require the pre-preparation of relevant multimodal data for training. 
As a consequence, preparing data for multimodal training demands substantial human and financial resources.
Thus, we ask: Do we have a simple and automated way to construct a large amount of multimodal data in specific domain?

\begin{figure*}[tb]
  \centering
  \includegraphics[width=0.9\textwidth]{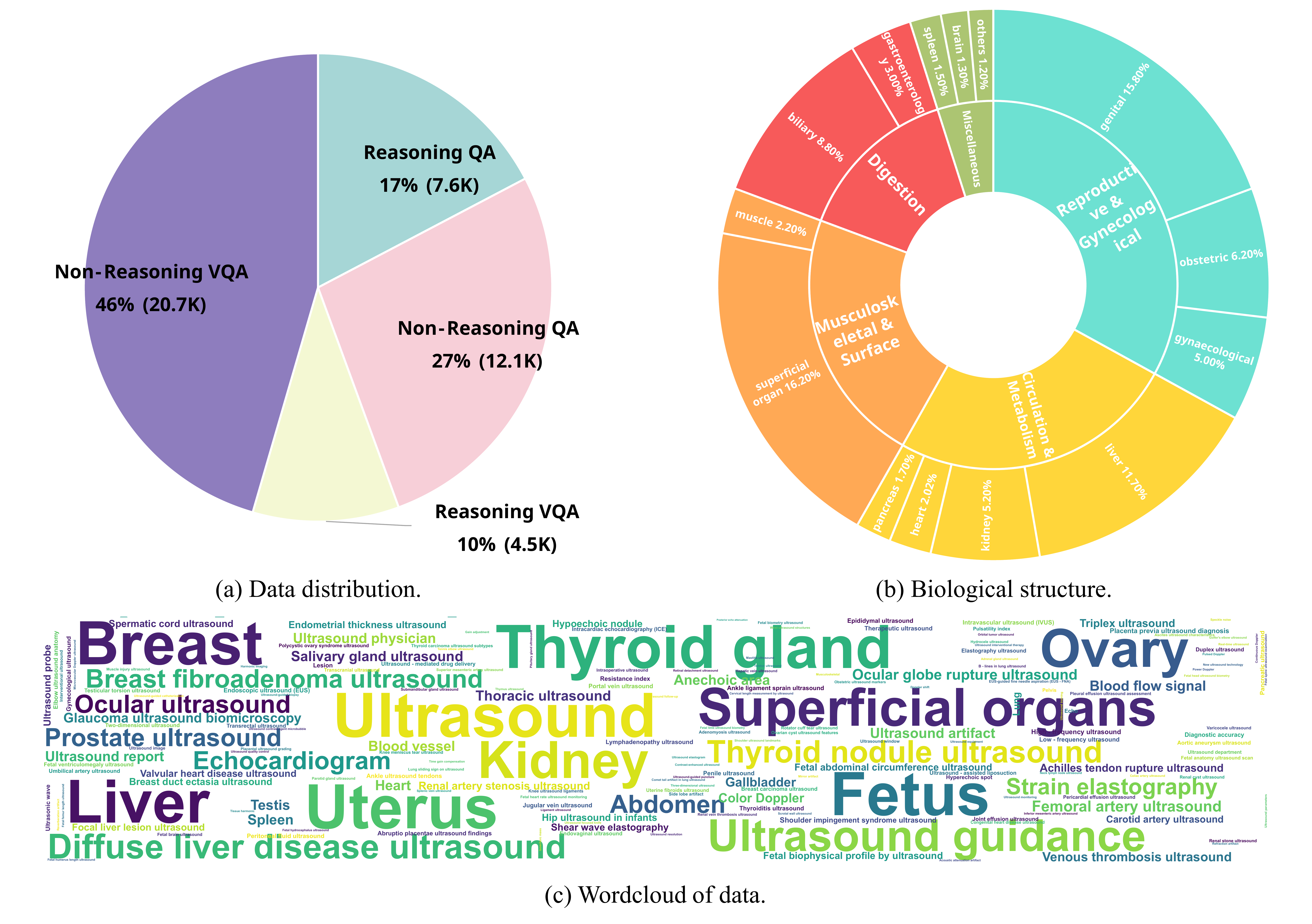}
  \caption{Statistical overview of ReMUD. ReMUD incorporates ultrasound knowledge across various anatomical regions, enabling the model to acquire comprehensive and extensive ultrasound expertise.}
  \label{fig:distribution}
\end{figure*}

In this paper, we propose a novel image-text data generation pipeline to create specific domain quadruplets (image, question, thinking trace and answer) from scratch. 
In a specific domain, there are a large number of professional books and papers, which are often stored in the form of PDF. 
These materials contain a wealth of curated image-text information. 
The images along with their corresponding text descriptions are scattered throughout these resources. 
Taking the medical ultrasound domain as an instance, we utilized the bounding box function of Qwen2.5-VL~\cite{bai2025qwen25vltechnicalreport} and the multimodal image-text recognition and generation capacities of GPT-4o, Gemini-2.0-Flash-Thinking-Exp~\cite{gemini20_flash_thinking} to establish a multimodal dataset \textbf{ReMUD} specific to the ultrasound domain. 
Subsequently, by performing supervised fine-tuning with Qwen2.5-VL-7B-Instruct serving as the base model, the fine-tuned model ReMUD-7B was developed. This model, equipped with reasoning ability, performs better than general-domain multimodal large language models on the test dataset USTQ-Knowledge and UVQA-Diagnosis we established.
Specifically, our paper makes the following contributions:
\begin{itemize}
    \item \textbf{Specific domain quadruplets generation pipeline.} We present a novel data generation pipeline to create domain specific quadruplets (image, question, thinking trace and answer) from domain specific materials which can ensure the accuracy of professional knowledge. This requires no manual annotations or pre-generated dataset and it is applicable to various general specific domains.
    \item \textbf{ReMUD.} We establish the first open-source medical ultrasound domain supervised fine-tuning (SFT) dataset ReMUD, which contains 45,000+ reasoning and non-reasoning SFT QA and VQA data (see \mbluetext{Fig.~\ref{fig:distribution}}). It incorporates text Question Answering data, Visual Question Answering data, reasoning data, and non-reasoning data. Two test datasets USTQ-Knowledge and UVQA-diagnos have been compiled to evaluate the ultrasound capabilities. The reasoning model ReMUD-7B, which is fine-tuned by ReMUD on Qwen2.5-VL-7B-Instruct, outperforms other general-domain MLLMs.
    \item \textbf{Open-source.} To facilitate research of specific domain and medical ultrasound field, we will release the following assets to the public: the ReMUD instruction-following dataset, the codebase for data generation and the parameters of ReMUD-7B. 
\end{itemize}

\section{Related work}

\begin{figure*}[tb]
  \centering
  \includegraphics[width=0.85\textwidth]{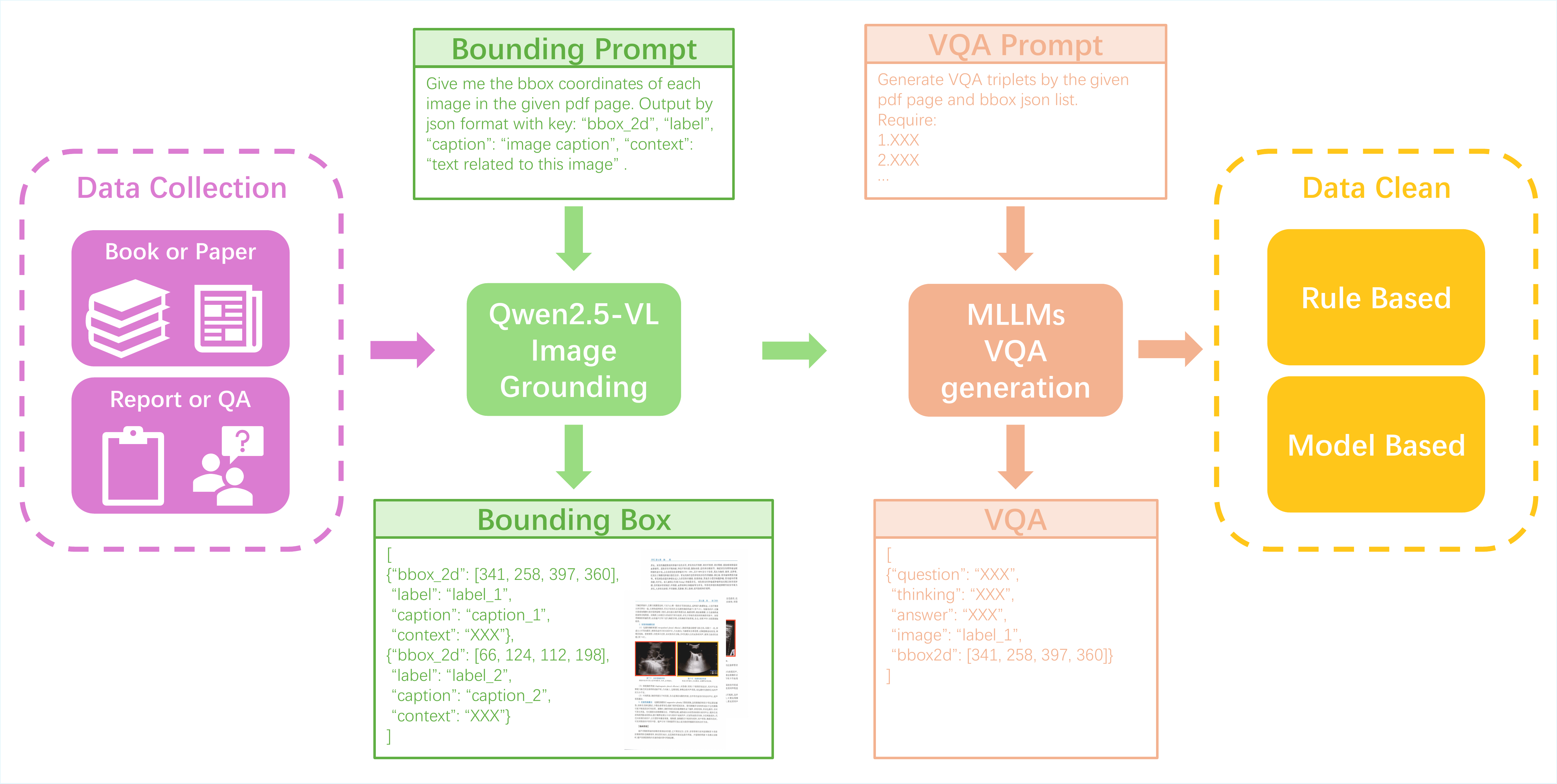}
  \caption{Flowchart illustrating the process of visual question answering (VQA) data generation, incorporating data collection, image grounding, VQA creation, and Data Cleaning.}
  \label{fig:workflow}
\end{figure*}

\subsection{Multimodal large language model}
Multimodal Large Language Models (MLLMs) have witnessed a surge in research recently~\cite{kimiteam2025kimivltechnicalreport, bai2025qwen25vltechnicalreport, wu2024deepseekvl2mixtureofexpertsvisionlanguagemodels, zhu2025internvl3exploringadvancedtraining, gemmateam2025gemma3technicalreport}. 
In architecture, they typically integrate pre-trained modality encoders (e.g., Vision Transformer visual encoder and its variants~\cite{dosovitskiy2020image, radford2021learningtransferablevisualmodels, liu2021swin}), pre-trained LLMs (such as LLaMA~\cite{touvron2023llamaopenefficientfoundation}, Vicuna~\cite{vicuna2023}, and Qwen series~\cite{qwen2025qwen25technicalreport, yang2024qwen2technicalreport}), and modality interfaces for multimodal interaction~\cite{li2023blip2bootstrappinglanguageimagepretraining}. 
Training MLLMs involves pre-training with large scale image-text data for modality alignment~\cite{liu2023visual}, instruction-tuning to enhance task generalization, and alignment tuning to match human preferences. 
Evaluation methods include closed-set evaluations on task-specific datasets and open-set assessments like manual scoring, model scoring, and case studies. Additionally, extended techniques such as Multimodal In-Context Learning (M-ICL)~\cite{doveh2024towards}, Multimodal Chain of Thought (M-CoT)~\cite{zhang2024multimodalchainofthoughtreasoninglanguage}, and LLM-aided visual reasoning~\cite{gupta2023visual} have been developed to enhance MLLMs' capabilities. 
However, challenges remain, including handling long-context multimodal information~\cite{zhang2024mm} and improving instruction-following~\cite{qian2024mia}, presenting opportunities for future research. In ultrasound field, LLaVA-Ultra~\cite{guo2024llavaultralargechineselanguage} is the first ultrasound multimodal large language model that trained by professional ultrasound multi-modal data, but its data and model parameters are not open-sourced.

\subsection{Visual Question Answering data generation}

Manually creating Visual Question Answering (VQA) datasets is a time-consuming and expensive process~\cite{rein2023gpqagraduatelevelgoogleproofqa}.
Template-based methods have been developed to address VQA data shortage. They generate QA pairs using templates, creating datasets like CLEVR~\cite{johnson2016clevrdiagnosticdatasetcompositional} and MIMIC-Diff-VQA~\cite{hu2023expert}. But models trained on such data perform poorly on complex human-written questions~\cite{shumailov2024ai}. 
Later, some studies use image descriptions and LLMs to generate questions and answer by using LLMs' high contextual understanding and reasoning ability to answer the question. For instance, Path-VQA~\cite{he2020pathvqa30000questionsmedical} is obtained by extracting pathology images and captions from publicly-available textbooks and digital libraries, using natural language processing techniques to generate question-answer pairs from the captions. Recently, there are even more MLLMs~\cite{openai2024gpt4ocard, gemini20_flash_thinking} to assist in the generation of VQA data.

\subsection{Test-time reasoning}
Test-time reasoning research has explored diverse methods to scale test-time compute. Parallel methods like majority voting~\cite{snell2024scalingllmtesttimecompute, brown2024large} and Best-of-N~\cite{levi2024simplemodelinferencescaling, irvine2023rewardingchatbotsrealworldengagement} generate multiple solutions simultaneously, yet may not fully tap into a model's reasoning. Sequential approaches allow for iterative refinement like Monte-Carlo Tree Search (MCTS)~\cite{gao2024interpretablecontrastivemontecarlo} and guided beam search~\cite{xie2023selfevaluationguidedbeamsearch}. OpenAI's o1 model~\cite{openai_learning_to_reason} spurred interest in test-time scaling, with replication efforts like DeepSeek-R1~\cite{deepseekai2025deepseekr1incentivizingreasoningcapability}, though openly replicating its scaling behavior was challenging. Other research focused on improving reasoning via continued training on specialized corpora, new training methodologies, and prompting techniques~\cite{yang2024syntheticcontinuedpretraining, yu2024metamathbootstrapmathematicalquestions, luo2025wizardmathempoweringmathematicalreasoning}.

\section{Ultrasound domain knowledge curation to create \textbf{ReMUD}}

\subsection{Data generation from  scratch}
In this section, we introduce our process for collecting a large scale multimodal data in medical ultrasound field and then creating \textbf{ReMUD} by automated image-text VQA generation. This is the first dataset for multimodal large language models specifically tailored to medical ultrasound. ReMUD is compiled from various publicly available ultrasound materials on the internet, such as professional ultrasound textbooks, clinical ultrasound guidelines, and public ultrasound datasets, covering a wealth of basic ultrasound knowledge. ReMUD is mainly in Chinese, with a small amount of English content.

\begin{figure*}[tb]
  \centering
  \includegraphics[width=0.8\textwidth]{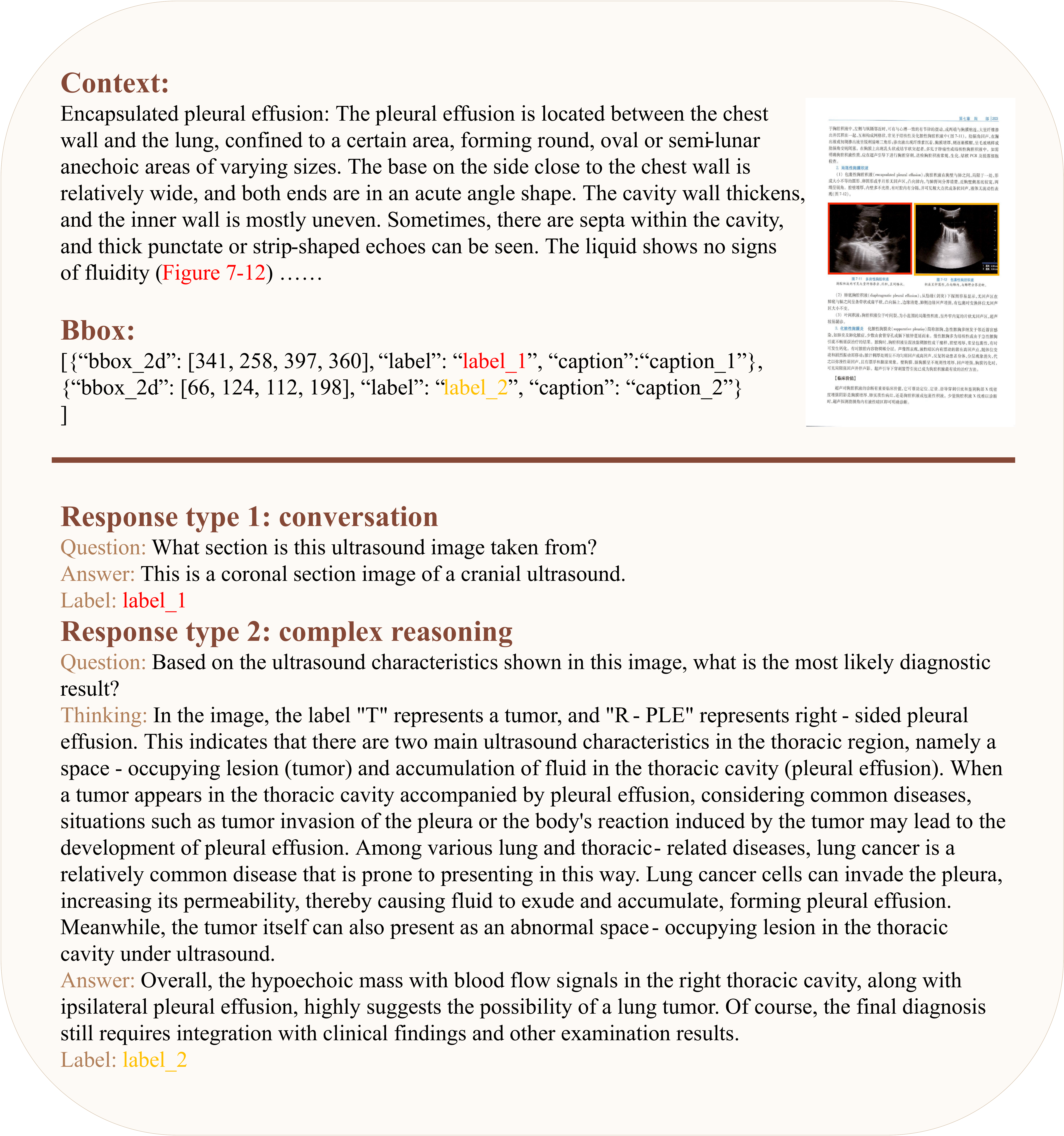}
  
  \caption{An example for VQA generation. By leveraging bounding box and page image, GPT-4o or Gemini-2.0-Flash-Thinking-Exp APIs can generate required data.}
  \label{fig:vqaexp}
\end{figure*}

We found that there are few publicly available ultrasound data that involve ultrasound multimodality, i.e., datasets with ultrasound image-text pairs. Therefore, we constructed the ultrasound dataset from scratch, e.g., classic ultrasound textbooks, up-to-date ultrasound guidelines, open source ultrasound dataset, and ultrasound paper, as a way to construct \textbf{ReMUD}. \mbluetext{Fig.~\ref{fig:workflow}} shows the general work flow of our approach. Specifically, we generate \textbf{ReMUD} with two categories.

\begin{subsubsection}{Text-only data}
Text-only data predominantly comprises textual ultrasound knowledge sourced from a diverse array of materials, including medical books, cutting-edge research papers, and in-depth technical documents. These materials serve as rich reservoirs of information, encompassing a wide spectrum of ultrasound concepts, ranging from fundamental principles to the latest advancements in the field. The data processing process is divided into three steps as follows:
\begin{itemize}
    \item First, we conducted an extensive crawl of publicly available data including medical books, cutting-edge research papers, and in-depth technical documents on the Internet covering a large amount of basic ultrasound knowledge, to strengthen the model's capability on ultrasound basics.
    \item Second, for formatted data, such as curated ultrasound knowledge question banks and widely recognized open-source ultrasound datasets, we adopted a structured conversion process. These data sources, despite already having a certain degree of organization, needed to be in a format that model could easily process. We applied specific rules to transform them into the JSON format. On the other hand, there were also unformatted data that presented unique challenges. These data were stored in PDF or image formats, which are not directly amenable to text-based analysis. To overcome this hurdle, we turned to Optical Character Recognition (OCR) methods~\cite{du2021ppocrv2bagtricksultra, li2022ppocrv3attemptsimprovementultra}. OCR technology is capable of converting the visual text within PDFs and images into machine-readable text. 
    \item After successfully converting the unformatted data into text format, we harnessed the power of the GPT-4o or Gemini-2.0-Flash-Thinking-Exp APIs. These advanced language model APIs were used to generate triplets consisting of a question, a thinking trace, and a generated answer. We set a strict prompt that requiring these generated triplets must be derived entirely from the textual information provided. This ensures that the generated content is relevant and reliable.
\end{itemize}
The questions and answers can take various forms, such as multiple-choice questions, which are useful for testing the model's knowledge recall and decision-making ability, or dialogues, which simulate real-world interactions and can better evaluate the model's understanding and in the ultrasound domain.
\end{subsubsection}

\begin{subsubsection}{Image-text data}
To create multimodal datasets, the acquisition and processing of image-text data  present a significant challenge. This is because generating high-quality image-text data usually demands a substantial amount of manual labor. Each image needs to be meticulously analyzed, and relevant text information, such as descriptions, captions, and annotations, must be accurately associated with it. This manual labeling work is not only time-consuming but also requires a high level of expertise. For instance, in a medical multimodal dataset, where images might be X-rays or MRIs, medical professionals are often needed to provide accurate text descriptions, which further complicates the process and increases the cost. 
Here we propose an automated annotation approach for the generation of image-text VQA data:
\begin{enumerate}
    \item Firstly, we use the precise object grounding function of Qwen2.5-VL to generate bounding boxes of images in each page of a PDF. Each bounding box corresponds to a piece of JSON data containing bounding box coordinates, labels and image caption. 
    \item Second, we take the pages with the generated bounding box labels and their corresponding JSON data and pass them to powerful multimodal Large Language Model APIs, such as GPT-4o or Gemini-2.0-Flash-Thinking-Exp~\cite{gemini20_flash_thinking}. Through the use of carefully crafted specific prompts, these APIs are able to generate Visual Question Answering (VQA) triplets related to each bounding box. These triplets consist of a question about the image content within the bounding box, a corresponding answer, and a thinking trace.
    \item Finally, to complete the establishment of a comprehensive image-text dataset, we utilize the bounding boxes to save the corresponding images. This step is essential for realizing the one-to-one correspondence between the image and the text data. By ensuring this correspondence, we can create a well-organized multimodal dataset where each image has its associated text information, including the bounding box details, labels, captions, and VQA triplets.
\end{enumerate}
From above, we get quadruplets each containing a image path, a question, a thinking trace and an answer (see \mbluetext{Fig.~\ref{fig:vqaexp}} for one of the few-shot examples.).

\begin{figure}[]
  \centering
  \includegraphics[width=0.8\columnwidth]{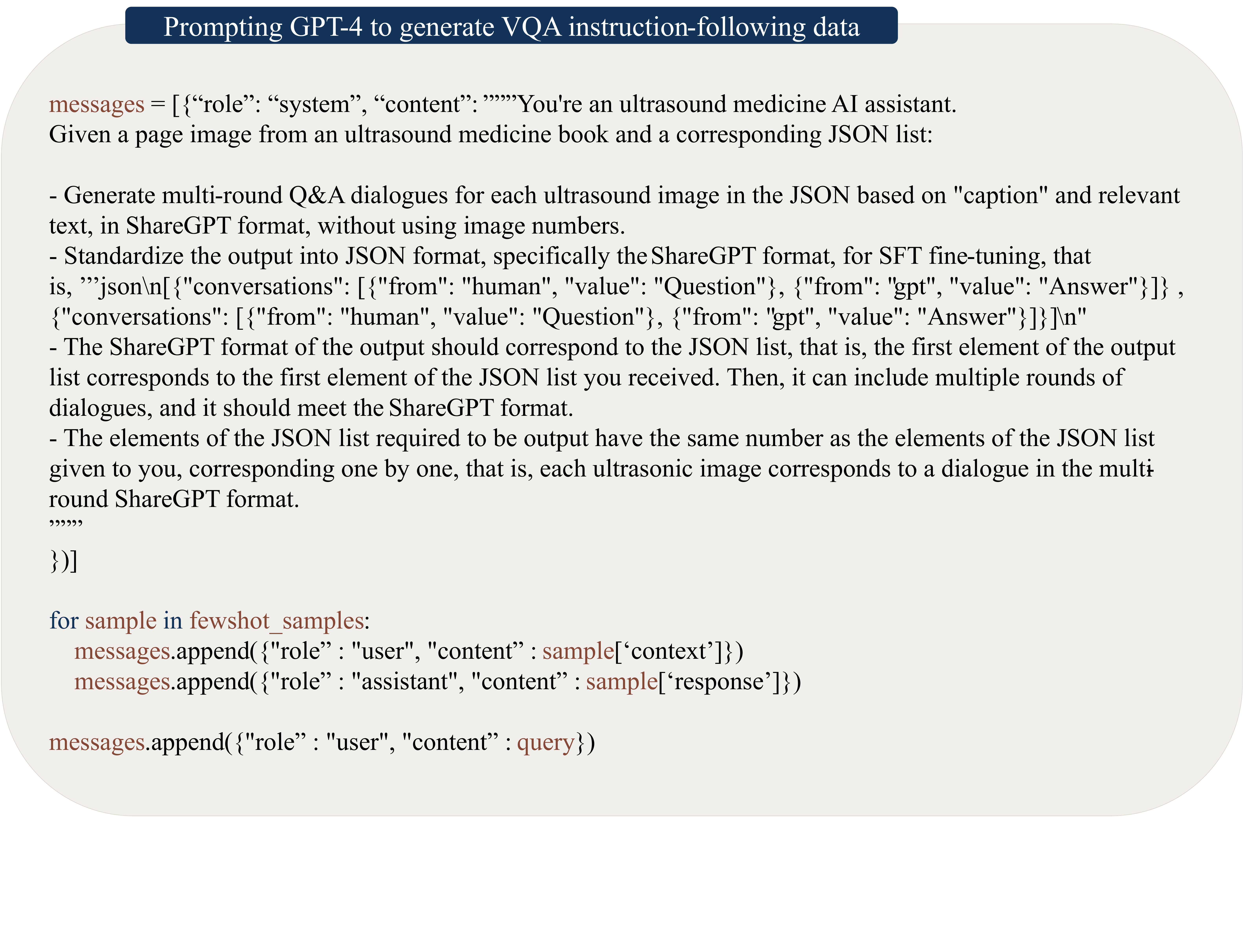}
  \caption{Messages we use to prompt GPT-4o to generate VQA instruction-following data. Different adjustments to the prompts may be made when handling different data. See Fig~\ref{fig:vqaexp} for one of the few-shot examples.}
  \label{fig:prompts}
\end{figure}

\end{subsubsection}

Similar to DeepSeek-R1~\cite{deepseekai2025deepseekr1incentivizingreasoningcapability}, we divide the dataset into reasoning-data and non-reasoning data. 
Specifically, for the generated factual QA dialogues, we do not generate the thinking trace, and for the generated multiple-choice questions or diagnostic type questions, we generate the thinking trace conform to the provided data.

\begin{figure*}[tb]
  \centering
  \includegraphics[width=0.618\textwidth]{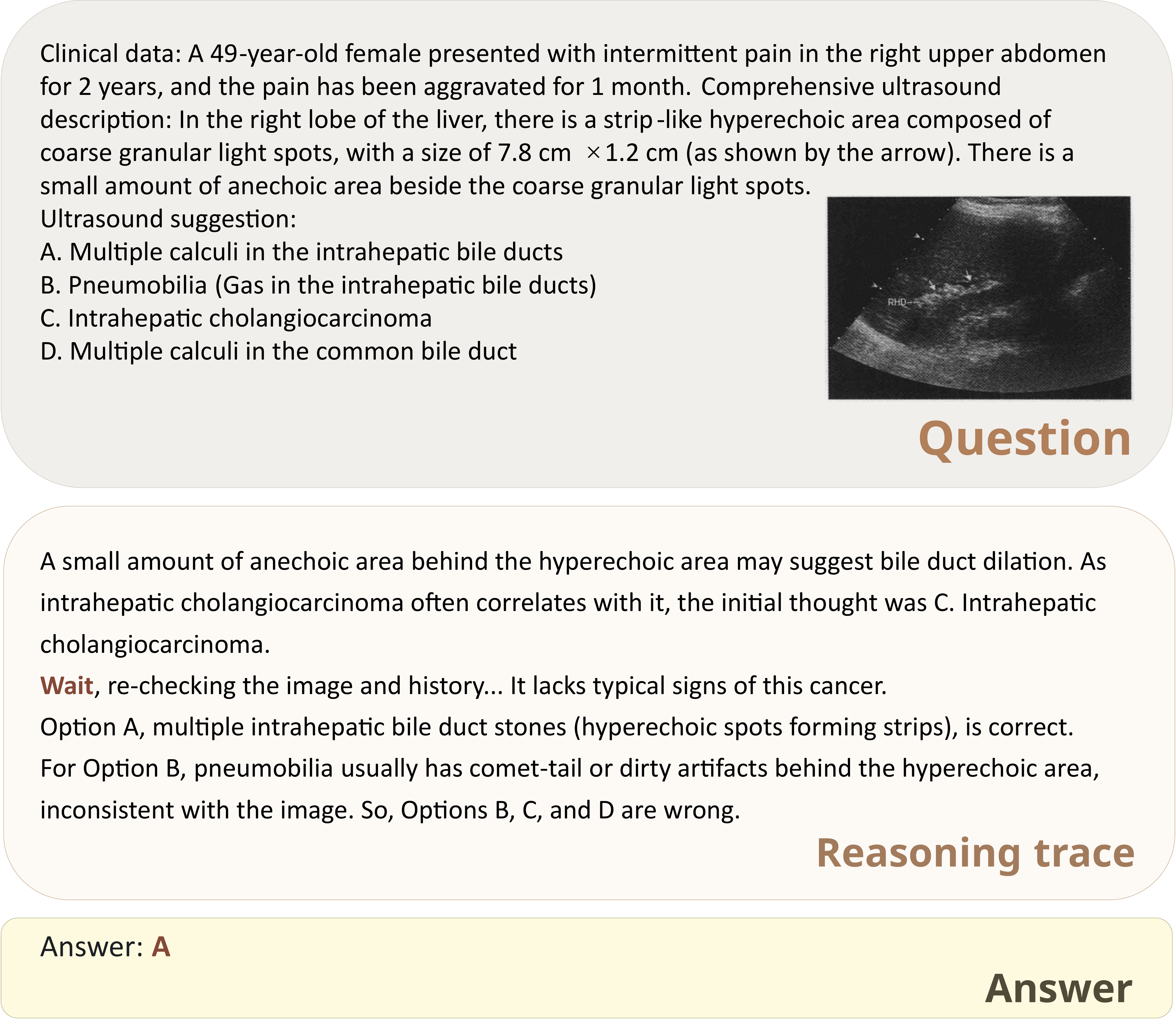}
  \caption{\textbf{Budget-forcing with Qwen2.5-VL-Ultra-7B}. Add the marker "Wait, " to the chain of thought to prompt the model to think again. After the model had incorrect thinking, it thought again and corrected the diagnostic result.}
  \label{fig:budgetforcing}
\end{figure*}

\subsection{Data cleansing and supplementation}
We produced a large number of text-only and image-text data through automated methods, but these data may contain errors and incompleteness, so we cleaned and supplemented the data according to the following three principles: Quality, Specialty, and Diversity.

\subsubsection{Quality}
To ensure data quality, we first remove incorrect data from API response errors and data with non-existent image addresses. Subsequently, we employ a pre-trained binary classification network to assess the legitimacy of images saved via bounding boxes. Finally, we utilize the Gemini-2.0-Flash-Thinking-Exp API to evaluate if the generated triplets are supported by page information and if the image-text match is of high quality through a scoring system, removing low-quality data to enhance the dataset's reliability for model training and evaluation.

\subsubsection{Professionalism}
In line with the expert advice of professional doctors, we meticulously curate professional ultrasound data by leveraging a diverse array of authoritative resources, including specialized professional books, cutting-edge research papers, and comprehensive clinical guidelines. When making API calls, we incorporate specific requirements within the prompt words. Specifically, we stipulate that the generated answers must be traceable back to the data we have provided. This ensures the accuracy and relevance of the output, maintaining a high standard of data quality. Moreover, when dealing with publicly accessible ultrasound multiple-choice question bank data that features only one correct answer, we adopt a multi-step verification process. First, we utilize the capabilities of GPT-4o to obtain its proposed answers and detailed explanations for each question. After that, we carefully sift through these responses, retaining only those answers and explanations that are correct. To further validate the retained content, we subject them to a rigorous review by professional doctors. This combination of AI-assisted analysis and expert human judgment helps us to create a reliable and high-quality ultrasound dataset for subsequent research and application.

\subsubsection{Diversity}
We collect data encompassing diverse ultrasound domains like breast, thyroid, and fetal ultrasound. Additionally, we gather publicly accessible ultrasound datasets from the Internet, such as those for breast nodule benign and malignant classifications and ultrasound report generation~\cite{li2024ultrasoundreportgenerationcrossmodality, chen2021usclpretrainingdeepultrasound}, and transform them into triplets for better utilization. Moreover, we filter relevant information from publicly available medical QA datasets, Path-VQA~\cite{he2020pathvqa30000questionsmedical} and PMC-OA~\cite{lin2023pmc}, based on ultrasound keywords and incorporate it into \textbf{ReMUD}, aiming to create a comprehensive and high-quality dataset for in-depth research and application development in the ultrasound field.

\begin{table}[]
\centering
\caption{ReMUD consists of four train dataset and two test dataset. Both of them are generated from open-sourced ultrasound materials.}
\label{tab:ReMUD}
\resizebox{\columnwidth}{!}{
\begin{tabular}{lcccc}
\hline
Dataset             & \multicolumn{1}{l}{Multimodality} & Thinking trace            & \multicolumn{1}{l}{Split} & \multicolumn{1}{l}{Count} \\ \hline
QA\_reasoning       & text                              & $\checkmark$ & train                     & 8,000+                     \\
QA\_non\_reasoning  & text                              & $\times$     & train                     & 20,000+                    \\
VQA\_reasoning      & image, text                       & $\checkmark$ & train                     & 4,000+                     \\
VQA\_non\_reasoning & image, text                       & $\times$     & train                     & 11,000+                    \\
USTQ-Knowledge      & text                              & $\checkmark$ & test                      & 371                       \\
UVQA-Diagnosis      & image, text                       & $\checkmark$ & test                      & 386                       \\ \hline
\end{tabular}%
}
\end{table}

\subsection{Final selection of \textbf{ReMUD}}

Through above methods, we get \textbf{ReMUD} containing text-only and image-text triplets. In order to evaluate the model's capability in ultrasound field, We have created two test datasets based on different data sources: USTQ-Knowledge and UVQA-Diagonis. USTQ-Knowledge automatically generated from professional ultrasound books. UVQA-Diagnosis is compiled based on the publicly available ultrasound report dataset~\cite{li2024ultrasoundreportgenerationcrossmodality}.
Given the potential presence of redundant data, we utilize n-gram (n=12) analysis and sentence transformer to efficiently identify and eliminate duplicates. This step is crucial as it ensures the uniqueness and quality of the test data, which in turn leads to more reliable evaluation results. 
These test data provide a broader range of scenarios and challenges for the model to encounter.
Ultimately, \textbf{ReMUD} contains $45k+$ training data and $1k+$ test data with totally $20k+$ images (see \mbluetext{Tab.~\ref{tab:ReMUD}}). The test data are in the form of multiple-choice questions and are accompanied by thinking trace.
This structure allows us to not only evaluate the model's ability to select the correct answer but also understand the reasoning process it undertakes, thereby facilitating a more in-depth and comprehensive assessment of the model's capabilities in the ultrasound field.

\section{Method}
We used the Qwen2.5-VL-7B-Instruct model~\cite{bai2025qwen25vltechnicalreport} as our base model. On this basis, we performed supervised fine-tuning on it using the \textbf{ReMUD} train dataset, and employed the budget forcing~\cite{muennighoff2025s1simpletesttimescaling} method to enhance its reasoning ability.

\subsection{Budget forcing}
Inspired by the innovative approach of s1~\cite{muennighoff2025s1simpletesttimescaling}, we integrate the concept of budget forcing into our model's inference process. This strategic decision aims to optimize the model's performance and output quality. Budget forcing, in our context, focuses on regulating the length of the chain of thoughts, which is crucial for maintaining the model's efficiency and the coherence of its reasoning.

More specifically, we target the content enclosed within the <think></think> tags. These tags serve as markers for the model's internal thought process representation. By controlling the length of this content, we can ensure that the model's reasoning neither becomes too concise to be meaningful nor too verbose to be practical. When the content within these tags is too short, it might indicate that the model has not fully explored the question. In such cases, we introduce a "Wait" tag. This "Wait" tag acts as a trigger for the model to continue processing, allowing it to generate more in-depth and comprehensive output. Conversely, if the content is too long, it could lead to inefficiencies and potential over-thinking. To address this, once the content exceeds the predefined length limit, we truncate it. After truncation, we add the </think> tag as a closure to maintain the integrity of the tag structure. This ensures that the model's thought representation remains well-formed and can be properly processed in subsequent operations. See \mbluetext{Fig.~\ref{fig:budgetforcing}}.

\begin{table*}[tb]
\centering
\caption{Comparison between different MLLMs. ReMUD-7B achieves better results than other models. DeepSeek-R1 is a single-modality large language model, so it cannot be tested on UVQA-Diagnosis. UVQA-Diagnosis test dataset has three ultrasonic parts: breast, liver and thyroid. "w/o BF" means without using budget forcing. ReMUD-3B is the fine-tuned model based on Qwen2.5-VL-3B-Instruct. It shows that scaling law still holds.}
\label{tab:benchmark_comparison}
\resizebox{\textwidth}{!}{%
\begin{tabular}{lccccc}
\hline
                              &                                  & \multicolumn{4}{c}{UVQA-Diagnosis} \\ \cline{3-6} 
\multirow{-2}{*}{Model}       & \multirow{-2}{*}{USTQ-Knowledge} & breast  & liver  & thyroid  & all  \\ \hline
Qwen2.5-VL-7B-Instruct        & 63.1                             & 41.4    & 37.3   & 36.9     & 38.2 \\
GPT-4o                        & 78.3                             & 62.9    & 62.7   & 62.8     & 62.7 \\
Gemini-2.0-Flash-Thinking-Exp & 70.8                             & 68.3    & 66.2   & 54.9     & 62.4 \\
Gemini-2.5-Pro-Exp            & 78.1                             & 70.9    & 70.7   & 58.1     & 66.6 \\
DeepSeek-R1                   & 79.2                             & -       & -      & -        & Nan  \\
Claude-3.7-sonnet-thinking    & 66.8                             & 44.1    & 38.2   & 37.8     & 40.2 \\
ReMUD-3B & 71.2                             & 73.9    & 71.5   & 70.4     & 71.9 \\
ReMUD-7B w/o BF    & 78.5                             & 90.2    & 91.5   & 86.0     & 88.1 \\
\rowcolor[HTML]{DAE8FC} 
ReMUD-7B (ours)    & \textbf{80.1}                             & \textbf{91.8}    & \textbf{93.3}   & \textbf{88.4}     & \textbf{90.1 }\\ \hline
\end{tabular}%
}
\end{table*}

\subsection{Mixed supervised fine-tuning}
Different from LLaVA-Ultra~\cite{guo2024llavaultralargechineselanguage}, a ultrasound multimodal large language model that trained through ultrasound concept feature alignment, visual enhancement and adaptive sampling for data redundancy, we adopt DeepSeek-r1's distillation model training approach, which makes the training more efficiency and resource-saving.
\textbf{ReMUD} encompasses both reasoning and non-reasoning data, with reasoning data having <think></think> tags to represent the thinking trace.
Similar to DeepSeek-R1's distillation model and s1K, we fine-tune the model via supervised fine-tuning to enhance its performance, enabling non-reasoning dialogue capabilities and ultrasound disease diagnostic inference. Additionally, the trained loss function acts on the <think></think> tag to optimize the model's internal reasoning process for better handling of ultrasound-related diagnosis. 

\begin{figure*}[tb]
  \centering
  \includegraphics[width=0.8\textwidth]{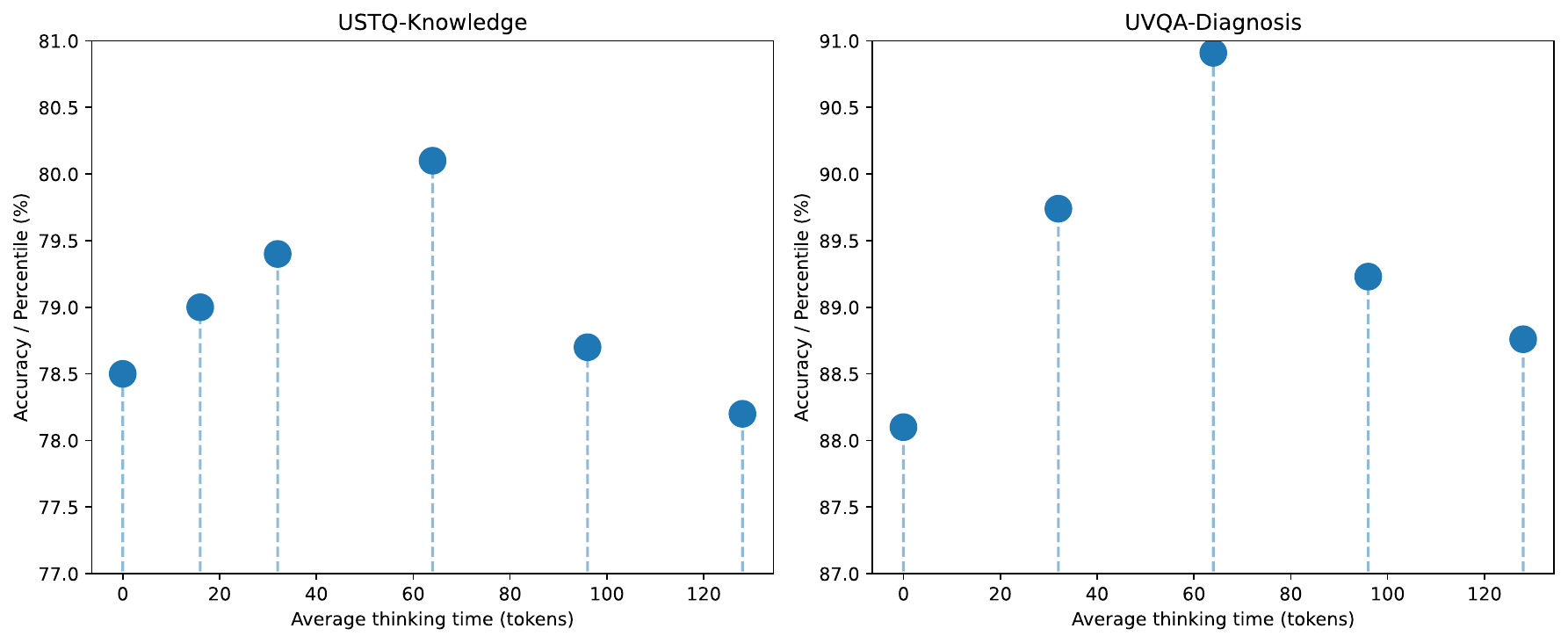}
  \caption{Test-time scaling with ReMUD-7B. We evaluate ReMUD-7B on UVQA-Diagnosis and USTQ-Knowledge using pass@1 and varying test-time compute length.}
  \label{fig:test_time_scaling}
\end{figure*}

\subsection{Evaluation metrics}
In order to make the test results of the model more stable, and in order to fully utilize the chain of thought to explore more possibilities, we chose the pass@1~\cite{chen2021evaluatinglargelanguagemodels} metric to evaluate our model:
\[
\text{pass@1} = \frac{1}{k} \sum_{i = 1}^{k} p_i,
\]
where $k$ denotes the total number of responses generated for each question, $p_i$ stands for the correctness of the $i$-th response, typically taking a binary value (1 for correct and 0 for incorrect). By summing these $p_i$ values over the range from $i = 1$ to $k$ and then dividing by $k$, we obtain the pass@1 value, which represents the average proportion of correct responses among the generated set. Our test dataset is in the form of multi-choice question, so it is easily to check the answer.

\section{Experiments}

\subsection{Setup}
We conduct supervised finetuning on the Qwen2.5-VL-7B-Instruct model leveraging ReMUD, and utilize LLaMA-Factory~\cite{zheng2024llamafactory} to train our model, ReMUD-7B. For model training, the learning rate is configured as 3e-5 with a cosine learning rate decay schedule, and the AdamW optimizer is employed. The entire finetuning procedure consists of $3$ epochs. All experiments are executed on a single NVIDIA A800 (80G) GPU with a warm-up rate set to the first 1\% of steps. The training process is completed in 4 hours on this single GPU. Related code can be found in \underline{\url{https://github.com/ShiDaizi/ReMUD}}.
\subsection{Test data}
Due to the lack of publicly available ultrasound benchmarks, we have created two different types of test datasets for ReMUD to evaluate the ultrasound-specific capabilities of MLLM models. The descriptions of the specific evaluation datasets are as follows:
\begin{itemize}
    \item \textbf{USTQ-Knowledge.} USTQ-Knowledge is automatically generated from ultrasound books, guidelines, and papers. It only contains text information. Specifically, ultrasound knowledge is sent to GPT-4o, which is required to generate text-based ultrasound multiple-choice questions based on the provided content, and the answers should be found within the given content. A total of $385$ questions were generated, and after manual screening, $371$ questions were obtained. 
    \item \textbf{UVQA-Diagnosis.} UVQA-Diagnosis is compiled based on the publicly available dataset, Chinese Ultrasound Report Dataset~\cite{li2024ultrasoundreportgenerationcrossmodality}, which contains $7k+$ ultrasound reports with $30k+$ images. By utilizing the information provided in the ultrasound reports, we have transformed the ultrasound diagnoses and the observed contents into the form of text-and-image single-choice questions, which are used to evaluate the model's capabilities in diagnosing ultrasound reports. UVQA-Diagnosis contains a total of 386 single-choice text-and-image questions.
\end{itemize}
\subsection{Evaluation}
We evaluate ReMUD-7B's ultrasound capability on USTQ-knowledge and UVQA-Diagonsis. The UVQA-Diagnosis test set can be further categorized into three anatomical regions according to the specific organs under diagnosis: thyroid, breast, and liver. We conducted separate validation tests for each of these three anatomical regions. 
Specifically, we set the temperature parameter at $0.6$ and the top-p parameter at $0.7$ to generate $k=4$ independent thinking-trace responses for each question in the test set. This configuration preserves controlled randomness in generation while maintaining a concentrated quality distribution of answers. We selected the pass@1 metric as the primary evaluation criterion, which measures the model's ability to produce at least one completely correct answer across multiple attempts—particularly suitable for medical diagnosis scenarios requiring precise reasoning.

For comparative baselines, we included state-of-the-art multimodal large language models: GPT-4o, Gemini-2.0-Flash-Thinking-Exp, Gemini-2.5-Pro-Exp, Claude-3.7-sonnet-thinking, and Qwen2.5-VL-7B-Instruct. As there are currently no suitable medical multimodal large language model APIs available for use, the comparisons conducted are all based on general-domain multimodal large language models.
All baseline models adopted identical sampling strategies and evaluation protocols to ensure fairness.

As shown in Table~\ref{tab:benchmark_comparison}, our model ReMUD-7B significantly outperformed all baselines across evaluation metrics. 
Notably, on the ultrasound diagnostic dataset UVQA-Diagnosis, ReMUD demonstrates significantly higher scores than general-purpose models. Additionally, ReMUD also achieves a certain edge on the text dataset USTQ-Knowledge. These results indicate that our automatically generated training data ReMUD effectively enhances the foundation model's capabilities in ultrasound-related professional tasks.

\subsection{Test-time scaling}

ReMUD-7B is a reasoning model. Inspired by s1~~\cite{muennighoff2025s1simpletesttimescaling}, we make a experiment by using budget forcing to validate test-time scaling. We forced the addition of the "wait" tag to the parts where the length of the thinking trace was insufficient, enabling the model to think again. It was found that compared to cases without thinking trace, the model's accuracy could be improved when a thinking trace was added. 
We found that appropriately increasing the length of the thinking trace can improve test accuracy, but when the length of the thinking trace is further extended, the accuracy will instead decrease. The specific results can be seen in \mbluetext{Fig.~\ref{fig:test_time_scaling}}.

\section{Discussion}
We constructed a multimodal reasoning dataset ReMUD by leveraging publicly available online data and documents, which encompasses rich basic ultrasound knowledge and diagnostic expertise. Experimental validation shows that fine-tuning the Qwen2.5-VL-7B-Instruct model with ReMUD enables it to surpass current mainstream general multimodal large language models in ultrasound capabilities, demonstrating that this workflow and approach can effectively enhance the performance of small models in specialized domains. Unlike complex methods such as LLaVA-Ultra~\cite{guo2024llavaultralargechineselanguage} that require modal alignment and visual enhancement, we only adopt fine-tuning to endow the model with certain ultrasound knowledge and reasoning capabilities in a simple way (only requiring a single A800 GPU and half a day of training time). Moreover, our data construction process is not limited to ultrasound but is more applicable to more specific domains, as long as the corresponding data or document can be obtained. 

However, limited by the data acquisition method, we cannot obtain particularly high-quality image information, and the clarity of the images may not be sufficient. Meanwhile, data such as videos are also difficult to process, and we are still stuck in the generation of text-image data. 

Additionally, due to the lack of suitable medical multimodal large language model APIs at present, we are unable to conduct further comparative experiments. However, we will publicly release our data, models, and code to promote the development of medical multimodal large language models.

\section{Conclusion}
In this study, we successfully developed a pipeline for generating specific domain image-text data, which is crucial for enhancing the performance of MLLMs in specialized fields. By leveraging professional books and papers in the medical ultrasound domain, we created the ReMUD dataset, which significantly improves the performance of the fine-tuned ReMUD-7B model in medical ultrasound tasks compared to general-domain MLLMs. The open-sourcing of our dataset, codebase, and model parameters will promote further research in specific domain and medical ultrasound fields. Future work could explore applying this pipeline to more specific domains, improving the quality of generated data, and enhancing the reasoning capabilities of models. Additionally, more efficient test-time scaling methods can be investigated to optimize the performance of models during inference.

\bibliographystyle{ieeetr}
\bibliography{libarary.bib}

\end{document}